\pdfoutput=1

\documentclass[11pt]{article}

\usepackage[dvipsnames,rgb]{xcolor}
\usepackage[]{ACL2023}

\usepackage{times}
\usepackage{latexsym}
\usepackage{adjustbox}

\usepackage[T1]{fontenc}

\usepackage[utf8]{inputenc}

\usepackage{microtype}

\usepackage{inconsolata}
\usepackage{lipsum}
\usepackage{float}
\usepackage{amsmath}
\usepackage{algorithm}
\usepackage{algpseudocode}
\usepackage{booktabs}
\usepackage{cleveref}

\usepackage{listings}
\usepackage{tikz}
\usetikzlibrary{positioning}
\usepackage{tikzsymbols}
\usetikzlibrary{shapes.misc, positioning, arrows.meta, shapes.geometric}
\usepackage{roboto} 

\makeatletter
\tikzset{
    database/.style={
        path picture={
            \draw (0, 1.5*\database@segmentheight) circle [x radius=\database@radius,y radius=\database@aspectratio*\database@radius];
            \draw (-\database@radius, 0.5*\database@segmentheight) arc [start angle=180,end angle=360,x radius=\database@radius, y radius=\database@aspectratio*\database@radius];
            \draw (-\database@radius,-0.5*\database@segmentheight) arc [start angle=180,end angle=360,x radius=\database@radius, y radius=\database@aspectratio*\database@radius];
            \draw (-\database@radius,1.5*\database@segmentheight) -- ++(0,-3*\database@segmentheight) arc [start angle=180,end angle=360,x radius=\database@radius, y radius=\database@aspectratio*\database@radius] -- ++(0,3*\database@segmentheight);
        },
        minimum width=2*\database@radius + \pgflinewidth,
        minimum height=3*\database@segmentheight + 2*\database@aspectratio*\database@radius + \pgflinewidth,
    },
    database segment height/.store in=\database@segmentheight,
    database radius/.store in=\database@radius,
    database aspect ratio/.store in=\database@aspectratio,
    database segment height=0.1cm,
    database radius=0.25cm,
    database aspect ratio=0.35,
}
\makeatother

\lstdefinestyle{SQLStyle}{
    language=SQL,
    basicstyle=\ttfamily\scriptsize,
    backgroundcolor=\color{gray!10},
    frame=single,
    frameround=fttf,
    columns=fullflexible,
    breaklines=true,
    breakatwhitespace=true,
}

\lstdefinelanguage{CustomGrammar}
{
    sensitive=true,
    morekeywords={root, list_the, whats_the, suffix, prop_countable, prop_number, teacher_multi_cond, teacher_cond_taught, class_level_single, teacher_cond_age, class, class_avg, class_year, subject, single_subject, none, TABLES, CONDITIONS, WANT, ADDER_PROPS},
    morecomment=[l]{\#},
}

\title{SecureLLM: Using Compositionality to Build\\
Provably Secure Language Models for Private, Sensitive, and Secret Data}

\author{Abdulrahman Alabdulkareem*\and Christian M Arnold*$^{\dag}$\\ {\bf Yerim Lee} \and {\bf Pieter M Feenstra} \\ {\bf Boris Katz} \and {\bf Andrei Barbu}\\
  CSAIL and CBMM, MIT \\
  * Equal Contribution ${}^{\dag}$ US Space Force \\
  \texttt{\{arkareem,cmarnold,boris,abarbu\}@mit.edu}}

\begin{document}
\maketitle
\begin{abstract}
Traditional security mechanisms isolate resources from users who should not access them. We reflect the compositional nature of such security mechanisms back into the structure of LLMs to build a provably secure LLM; that we term SecureLLM. Other approaches to LLM safety attempt to protect against bad actors or bad outcomes, but can only do so to an extent making them inappropriate for sensitive data. SecureLLM blends access security with fine-tuning methods. Each data silo has associated with it a separate fine-tuning and a user has access only to the collection of fine-tunings that they have permission for. The model must then perform on compositional tasks at the intersection of those data silos with the combination of those individual fine-tunings. While applicable to any task like document QA or making API calls, in this work we concern ourselves with models that learn the layouts of new SQL databases to provide natural-language-to-SQL translation capabilities. Existing fine-tuning composition methods fail in this challenging environment, as they are not well-equipped for handling compositional tasks. Compositionality remains a challenge for LLMs. We contribute both a difficult new compositional natural-language-to-SQL translation task and a new perspective on LLM security that allows models to be deployed to secure environments today.
\end{abstract}

\section{Introduction}

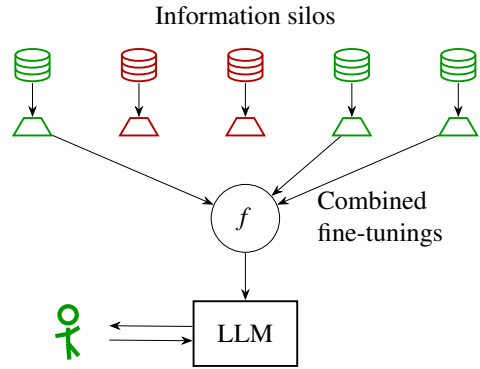
\begin{figure}[t]
\centering
\scalebox{0.9}{
  \begin{tikzpicture}[%
    user/.style={draw, rectangle, minimum size=1cm},
    func/.style={draw, circle, minimum size=1cm},
    arrow/.style={-{Stealth[]}, rounded corners}]

\node[OliveGreen, thick, database] (db1) at (0,0) {};
\node[BrickRed, thick, database, right=of db1] (db2) {};
\node[BrickRed, thick, database, right=of db2] (db3) {};
\node[OliveGreen, thick, database, right=of db3] (db4) {};
\node[OliveGreen, thick, database, right=of db4] (db5) {};

\node[draw, thick, OliveGreen, trapezium, below=0.5 of db1] (ft1) {};
\node[draw, thick, BrickRed, trapezium, below=0.5 of db2] (ft2) {};
\node[draw, thick, BrickRed, trapezium, below=0.5 of db3] (ft3) {};
\node[draw, thick, OliveGreen, trapezium, below=0.5 of db4] (ft4) {};
\node[draw, thick, OliveGreen, trapezium, below=0.5 of db5] (ft5) {};

\node[above=0.2 of db3] (silos) {Information silos};

\node[OliveGreen, below left=2.85cm and -1.4cm of db1, scale=3] (user) {\Strichmaxerl[1][10][30][40][4]};

\node[func, below=0.7 of ft3] (fnode) {\textit{f}};

\node[right=0.2 of fnode] (silos) {\begin{tabular}{l}Combined\\fine-tunings\end{tabular}};

\draw[arrow] (ft1) -- (fnode);
\draw[arrow] (ft4) -- (fnode);
\draw[arrow] (ft5) -- (fnode);

\draw[arrow] (db1) -- (ft1);
\draw[arrow] (db2) -- (ft2);
\draw[arrow] (db3) -- (ft3);
\draw[arrow] (db4) -- (ft4);
\draw[arrow] (db5) -- (ft5);

\node[draw, rectangle, thick, below=0.7 of fnode, inner sep=10pt] (LLM) {LLM};

\draw[arrow] ([yshift=-3pt]user.east) -- ([yshift=-3pt]LLM.west);
\draw[arrow] ([yshift=3pt]LLM.west) -- ([yshift=3pt]user.east);

\draw[arrow] (fnode) -- (LLM);
\end{tikzpicture}
}
\caption{No user-LLM interaction can access information silos that the user does
  not have permissions for; the fine-tunings associated with those silos are not
  connected to the user's LLM. This is the key component of SecureLLM: reflect
  the compositional nature of access security back into the structure of the
  fine-tunings of the LLM. LLMs have great difficulty generalizing to
  this composition environment. Worse, existing fine-tuning methods fail
  in this challenging compositional environment. We introduce new
  fine-tuning methods that perform well because they are selected specifically for this domain.}
\label{fig:main}
\end{figure}

Using language models in environments with sensitive information is fraught with problems. Models can be convinced to reveal information they should not, to answer questions they should not, to reveal their training data and prompts, to run API calls they should not, etc. Various approaches have tried to grapple with this problem, yet no prior work offers a method that provides any guarantees, merely offering mitigations. When dealing with sensitive data, this is practically and often legally insufficient. We provide the first method to build provably secure LLMs by reflecting the compositionality inherent in access security into the LLM.

We consider the scenario where a number of silos of information are available (\cref{fig:main}). These may be APIs, databases, or collections of documents with sensitive information. Each user has access to some subset of these silos. Nominally, one could imagine an exponential number of LLMs with an LLM for each subset of silos. This would guarantee security, as an LLM being used would only be trained on data that the user has access to. Creating and maintaining an exponentially-increasing number of models is impractical. Instead, we show how to achieve the same goals with a linear number of LLM fine-tunings. While using only one fine tuned model per silo, we can configure and compose a model specific to the user's permissions at runtime.

Knowledge about the target domain is critical. A silo may contain a large volume of documents, databases, or APIs, each with their own internal logic. Even Retrieval-Augmented Generation (RAG) is insufficient for such applications because it presupposes that the model understands the logic of a target domain. You cannot accurately retrieve something that you don't understand at all; imagine a new technology or capability. SQL translation tasks offer an extreme version of this where the target database schema must be known. Often, the schema itself is highly sensitive and should not be revealed to all users. This is the scenario we consider here.

To build such secure LLMs we rely on fine-tuning. For each information silo we fine-tune a model using a method that isolates the weight updates to a small part of the model which can then be excised and stored separately. Then, when a user interacts with the system, their permissions automatically determine the collection of fine-tunings that must be applied to the model and how those fine-tunings must be applied. For example, a user with access to silos A, B, and C that wants to communicate to another user with access to only A and B, would set up their model with positive fine-tunings for A and B in order to actively avoid topics related to C.

This idea is reminiscent of recent work like LoraHub \citep{compose_lorahub_huang2023} which also composes fine-tunings. Given a target task, LoraHub selects a set of fine-tunings, Low Rank Adapters (LoRAs) \citep{lora_hu2021}, that are added together. However, LoraHub is designed for soft tasks, where a model already tends to perform well, and where LoraHub's aim is to increase performance by a few percent. The domain we consider here is radically different, we seek compositionality for unrelated silos of information. Questions that require access to silos A and B, by definition cannot be answered with access to only A or only B. The underlying performance of models is nearly zero. LoraHub is not well-suited to such tasks and performs very poorly. It is only through exploiting the compositionality of security as a domain that our methods achieve much higher performance.

Our contributions are:
1. formulating a difficult new compositional task that LLMs have great difficulty with -- natural-language-to-SQL where not only is a model trained on queries of individual databases but also where the generated response requires cross-database joins,
2. formulating the notion of access security in terms of this task,
3. demonstrating that existing fine-tuning methods fail in this compositional environment,
4. the introduction of new compositional fine-tuning methods for this problem,
While we only concern ourselves with the task of understanding queries by translating them to SQL, our methods are generic and can be applied to numerous other domains like translating commands to API calls and answering questions from large collections of documents.

\section{Related Work}

\textbf{Model Composition}.  Our framework relies on composing LLM fine-tunings at inference time which follows a set of previous works that use model composition. A recent method combines pretrained LLM prompts each tuned for separate tasks to achieve generalization on downstream tasks \citep{compose_sun-etal-2023} which requires training at inference time. AdapterSoup composes fine-tunings by linearly averaging their weights depending on a criteria to determine which fine-tunings are relevant to the new domain \citep{compose_chronopoulou2023}. PEM Addition is a method that doesn't require further training such as composing fine-tunings using arithmetic operations directly on the weights \citep{compose_zhang2023}.
LoraHub is a recent simple framework that also composes different LoRa fine-tunings \citep{lora_hu2021} at inference time where each fine-tuning is trained on a different task \citep{compose_lorahub_huang2023}; we include comparisons using LoraHub and PEM Addition.

\textbf{Privacy Attacks}.  Many recent works discuss a range of different privacy attacks against large language models and Deep Learning models in general. Membership inference attacks are a type of privacy attack which try to determine if a piece of text was contained in the training data of a model possibly without access to the weights \citep{leak_hisamoto2020, leak_nasr2019, leak_hu2022}. An even larger security risk is posed by training data extraction attacks where large language models leak text in their training data verbatim \citep{leak_carlini2019} including personally identifiable information \citep{leak_inan2021}. This attack is shown to be successful even when such data was only mentioned in a single document and this behavior worsens with an increase in model size \citep{leak_carlini2021}. Similarly, training data extract attacks were effective on models fine-tuned on a smaller dataset \citep{leak_zanella2020}. With recent work tackling these privacy issues for Retrieval-Augmented Generation using multi party communication \citep{prag_zyskind2023}.

\textbf{Differential Privacy}.  A popular algorithmic technique to train machine learning models with certain privacy guarantees is differential privacy \citep{diffpriv_abadi2016deep} which has also been applied to large recurrent language models \citep{diffpriv_rnn_mcmahan2017}. Multiple recent works manage to use differentially private learning on large language models with hundreds of millions of parameters to achieve efficient differentially private fine-tuning with slight degradation in performance \citep{diffpriv_li2021, diffpriv_llm_yu2021}. Many other methods borrow inspiration from differential privacy like Confidentially Redacted Training which provably prevents memorization of the training data \citep{diffpriv_llm_zhao2022}. However, there are differences between Differential Privacy and our approach. In differential privacy, there exists a non-zero amount of privacy loss parameterized by the privacy budget ($\epsilon$ and $\delta$) from the resulting model as the privatization step minimizes but does not completely ensure that the updated model parameters do not leak private information. Additionally, there is a difference in what is considered private information compared to what is non-confidential. Differential Privacy considers the individual records within the training data as private such that any individual record is sufficiently obfuscated while holistic trends are still learnt by the resulting model. Comparatively, our notion of privacy ensures that every individual record in a private silo needs to be completely private including any and all holistic information gained from the silo.

A common trend in privacy preserving research into LLMs is the focus on preventing models from memorizing and/or leaking individual records or pieces of information while attempting to maintain a holistic distillation of the training data into the model. This distinguishes our work as we consider any and all information gleaned from a silo to be confidential no matter how granular or holistic it is. We also avoid falling back to probabilistic claims or claims in expectation and ensure that all claims of confidentiality are provably correct as the resulting model weights that an individual user interacts with have only been optimized using data that the user is authorized to access.

\section{Framework}

SecureLLM takes several fine-tunings each trained on distinct information silos and composes them at inference time. The goal of the composed model is to answer questions about both individual silos and questions that span silos. For example, in our case, a natural-language to SQL LLM would need to be able to generate joins across the databases of multiple silos to answer complex questions that have never been seen at training time. This is a trivial task for humans, but one that challenges LLMs. We go a step further: not only must such an LLM work, it must operate through a combination of fine-tunings, i.e., not only has it never seen combinations of silos at training time, its fine tunings have only ever seen a single silo each. This challenges, and defeats, current fine-tuning methods. The upshot of this difficult task is that it solves several key security problems for LLMs.

Given $N$ data silos $\{S_1, S_2, \cdots, S_N\}$ and $N$ fine-tuned LLMs $\{M_1, M_2, \cdots, M_N\}$ where $M_i$ has been fine-tuned on the data silo $S_i$, and given a set of target indices $T\subseteq\{1,2,\cdots,N\}$, the goal is to obtain a composed model $M_{T} := M_{T_1} \oplus \cdots \oplus M_{T_{|T|}}$ at inference time with no additional training such that $M_{T}$ is able to correctly answer any question about the information contained in the target silos $S_i, \forall i \in T$ and should fail to answer any question about information not contained in the target silos $S_j, \forall j \notin T$ as to not leak any information that the desired model $M_{T}$ is not intended to have. Additionally, the target model $M_{T}$ should be able to answer new \textit{union questions} $q_{union,ij} \in S_{i \cup j}$ where $i\in T \wedge j \in T$ where the question relies on information contained in both $S_i$ and $S_j$. We note that the union questions $q_{union,ij}$ are not answerable by any individual data silos, thus none of the individual models $M_i$ are able to answer any union questions while a successfully composed model should be able to answer such questions without the need of any training.

It is critical that the composed model $M_T$ has no knowledge of information silo that the user is not authorized to access, i.e. data silos $S_i, i \notin T$. Without this condition, a trivial solution is to train a single model $M_{All}$ on all data silos $\{1,\cdots,N\}$ however this approach is susceptible to leaking confidential information as the model would have knowledge of information contained in silos that users are not authorized to view and thus is not a valid approach. This approach is also problematic for scenarios that employ security through contradiction, in that some silos may directly contradict information in another silo in order to protect sensitive information (SecureLLM could potentially solve this by applying weights to silos of higher confidentiality). We refer to $M_{All}$ as the Exponential Model that has seen every combination and such a model is used as an insecure upper bound to performance in our experiments. 

An alternative to composing fine-tunings while also preserving privacy would be to create an exponential number of models, one for the powerset of information silos. This would maximize performance and minimize the amount of generalization needed, as long as one had a way to automatically generate cross-silo questions, perhaps with another LLM. This is obviously impractical. In essence, our method provides the advantages of the exponential approach but with linear storage and training runtime.

\subsection{Security as a relaxation}

What sets security-related issues apart from other fine-tuning combination scenarios is a relationship to a simple and easy-to-formalize idealized problem. Assume that each silo has a disjoint alphabet, that the contents of each silo is modeled by a state machine, and that the output the LLM must produce should be an utterance that is accepted by the union of the per-silo state machines. In this setting, security is simple. The optimal output depends on one silo at a time and there is no need to mix information about silos. The methods we provide here are optimal in this idealized setting because of its likeness to SQL.

For SQL generation tasks the model primarily must insert table and column names which are part of individual silos while following the overall structure of SQL which is silo-independent. Of course, real security scenarios are relaxations of this problem. For example, cross-silo question answering, is faithful to this idealized problem if portions of the answers are derived entirely from individual silos, while if they require cross-silo reasoning then QA becomes a relaxation. Current methods, as we show below, are not well-suited for this idealized problem or its relaxations. Formalizing this setting will hopefully lead to new composition operators.

\subsection{Composing fine-tunings}

We discuss several existing methods, none of which perform well. A few other plausible methods that also do not perform well are shown in the appendix. Finally we describe two new methods that do perform well with one clear winner. 

\textbf{LoraHub}\label{subsub:LoraHub} The LoraHub method for composition introduced by \citet{compose_lorahub_huang2023} is a two-step process involving element-wise summation of LoRA fine-tunings (\textit{COMPOSE}), and then learning weight optimizations via gradient-free methods to apply to each fine-tuning (\textit{ADAPT}). For this paper, we do not implement the \textit{ADAPT} stage because weights for every possible combination would need to be learned, and we could not say that this process is completed at inference time.



We observe that LoraHub performs poorly on the secure composition task. The authors warn that combining too many fine-tunings can lead to poor performance, however this cannot be the source of poor performance as we compose only up to three LoRAs.

\textbf{PEM Addition}\label{subsub:PEMAddition} The summation method introduced by \citet{compose_zhang2023} is similar to LoraHub, however, instead of summing the embeddings of the encoder and decoder prior to receiving the input $x$, one executes each fine-tuning independently at the attention-layer level, and then adds the result.  This version of summed composition shows improved performance over LoraHub.

\textbf{Average of Adapter Weights} Computing the simple average of each Lora fine-tuning response, as suggested by \citet{compose_chronopoulou2023}, $\sum_{i \to n}{\frac{L_i}{n}}$, produced compositions that were 50\% less effective than PEM Addition in initial informal tests.

\textbf{Variations of LogSumExp of Adapter Weights} \citet{du2020compositional} proposes a disjunctive composition process based on Energy Based Modeling, $-logsumexp(-E_1(x), -E_2(x), \cdots)$. Every variation tried performed significantly worse than PEM Addition, and upon closer inspection, this process substantially distorts encoder and decoder embeddings.

\textbf{Adapter Concatenation} The \citet{peft_library} library implements weight concatenation, however we found that concatenating LoRA encoder/decoder fine-tunings performed significantly worse than PEM Addition.

\subsection{Our Methods}

\textbf{Maximum Difference}\label{subsub:MaximumDifference} The intuition behind this method is to select the embeddings from each fine-tuning with the strongest response (either positive or negative) at each attention layer. In order to accomplish this, each LoRA fine-tuning is evaluated separately on input $x$. Then a mask of zeros with the same dimension as the output is created, $h_{max}$, to aggregate LoRA responses. For each LoRA fine-tuning response $L_i$, an element-wise comparison is made, and if the absolute values of the fine-tuning response is greater than the aggregated response, then the signed response from that fine-tuning replaces the element in the aggregated response. 

\textbf{Logit Composition}\label{subsub:LogitComposition} Given fine-tunings to compose $M_1, \cdots, M_n$ and input $x$, we define logit composition as performing the complete forward pass for each fine-tuning independently to obtain logit probabilities. We select the maximum value of each logit. One could instead sum logits for each fine-tuning. We found little difference between the two implementations, although the sum may have issues as the number of fine-tunings increases. 

Note that we are not claiming this method to be a superior compositional approach in every case. The requirements of compositional security are different than those of some other compositional tasks. In a sense, by its very nature, compositional security implies that most of the time every silo but one is irrelevant and confused, and one silo is likely to produce confident results. This motivates our compositional methods and explains why other methods preform so poorly.

\section{Data generation}

Our goal is to automatically create a challenging dataset for compositions of silos. There are countless other natural-language-to-SQL datasets out there, which we do not aim to replace. As such, we focus specifically on within and between silo questions. We also cover only a portion of SQL. We do not aim to exhaustively test how well models understand SQL, we aim to understand how well models generalize their knowledge from questions about individual silos to questions that span silos. As such, we consider only a subset of SQL which is otherwise an imposingly complex language.

We automatically generate SQL databases, one per silo, with 2-3 tables per database, that share columns which can be joined together both within and between databases. Databases are otherwise disjoint and on different topics. For each database we generate natural language questions along their equivalent SQL. Then, we generate questions and SQL pairs that span pairs and triples of databases. Two methods are used to generate these pairs: a CFG (see \cref{fig:union_example}) and ChatGPT 4 (see \cref{fig:gpt_obf_example}). The CFG generates both the SQL and the question in parallel. We do this at large scale, with 100,000 pairs per silo or combination of silos. To ensure that our results scale to more realistic queries we also generate 300 pairs per silo or combination of silos.

\begin{figure}[h]
\centering
\begin{tikzpicture}
    \node[draw, text width=7cm, align=left, font=\fontfamily{Roboto}\selectfont\scriptsize] (q1) at (0,0) {Q: What's the average age of all teachers that are older than 72 or that taught art classes for 9th graders in the school.   Answer:};
    \node[text width=6.988cm, align=left, below=-0.394cm of q1] (s1) {
    \begin{lstlisting}[style=SQLStyle]
    SELECT AVG(instructors.teacher_age)
    FROM instructors INNER JOIN classes 
        ON instructors.teacher_id = classes.teacher_id
    WHERE instructors.teacher_age >= 72 
        OR classes.class_subject = 'art' AND classes.level = 9
    \end{lstlisting}};
    \node[text width=7cm, align=center, below=-0.3cm of s1, font=\fontfamily{Roboto}\selectfont\small] (l1) {(a) Sample from Silo 1 ($S_1$)};


    \node[draw, text width=7cm, align=left, below=0.2 cm of l1, font=\fontfamily{Roboto}\selectfont\scriptsize] (q2) {Q: What's the minimum height of all appliances in the inventory that are currently unavailable in stores located in NY, CA, or MA and with a rating higher than or equal to 2 stars.  Answer:};
    \node[text width=6.988cm, align=left, below=-0.394cm of q2] (s2) {
    \begin{lstlisting}[style=SQLStyle]
    SELECT MIN(inventory.height)
    FROM inventory INNER JOIN store ON 
        store.store_id = inventory.store_id
    WHERE inventory.available = 0 
        AND (store.location = 'NY' 
         OR store.location = 'CA' 
         OR store.location = 'MA') 
        AND store.star_rating >= 2
    \end{lstlisting}};
    \node[text width=7cm, align=center, below=-0.3cm of s2, font=\fontfamily{Roboto}\selectfont\small] (l2) {(b) Sample from Silo 2 ($S_2$)};


    \node[draw, text width=7cm, align=left, below=0.2 cm of l2, font=\fontfamily{Roboto}\selectfont\scriptsize] (q3) {Q: Provide the names of all managers located in TX and the names of all teachers that are younger than 86 and that taught english, sociology, or art classes that achieved a grade higher than 89 in the database.  Answer:};
    \node[text width=6.988cm, align=left, below=-0.394cm of q3] (s3) {
    \begin{lstlisting}[style=SQLStyle]
    SELECT store.name
    FROM classes
    INNER JOIN instructors ON instructors.teacher_id = classes.teacher_id
    INNER JOIN store ON store.name = instructors.name
    WHERE store.location = 'TX' 
     AND instructors.teacher_age <= 86 
     AND (classes.class_subject = 'english' OR classes.class_subject = 'sociology' OR classes.class_subject = 'art') 
     AND classes.grade >= 89
    \end{lstlisting}};
    \node[text width=7cm, align=center, below=-0.3cm of s3, font=\fontfamily{Roboto}\selectfont\small] (l3) {(c) Sample from Union Silo 1,2 ($S_{1 \cup 2}$)};

\end{tikzpicture}
\caption{Examples of input/output pairs of a question paired with the target SQL query which are unconditional samples from a Context-Free Grammar.}
\label{fig:union_example}
\end{figure}

\begin{figure}[h]
\centering
\begin{tikzpicture}
    \node[draw, text width=7cm, align=left, font=\fontfamily{Roboto}\selectfont\scriptsize] (q1) at (0,0) {Q: What is the average age of instructors who are aged 72 or older or teach art at level 9?   Answer:};
    \node[text width=6.988cm, align=left, below=-0.394cm of q1] (s1) {
    \begin{lstlisting}[style=SQLStyle]
    SELECT AVG(instructors.teacher_age)
    FROM instructors INNER JOIN classes 
        ON instructors.teacher_id = classes.teacher_id
    WHERE instructors.teacher_age >= 72 
        OR classes.class_subject = 'art' AND classes.level = 9
    \end{lstlisting}};
    \node[text width=7cm, align=center, below=-0.3cm of s1, font=\fontfamily{Roboto}\selectfont\small] (l1) {(a) Sample from Silo 1  ($S_1$) that was rephrased with ChatGPT};


    \node[draw, text width=7cm, align=left, below=0.2 cm of l1, font=\fontfamily{Roboto}\selectfont\scriptsize] (q2) {Q: What's the minimum height of all appliances in the inventory that are currently unavailable in stores located in NY, CA, or MA and with a rating higher than or equal to 2 stars.  Answer:};
    \node[text width=6.988cm, align=left, below=-0.394cm of q2] (s2) {
    \begin{lstlisting}[style=SQLStyle]
    SELECT MIN(inventory.sloth)
    FROM inventory INNER JOIN store ON
        store.bear = inventory.bear 
    WHERE inventory.pony = 0
        AND (store.alpaca = 'NY' 
         OR store.alpaca = 'CA'
         OR store.alpaca = 'MA')
        AND store.raccoon >= 2
    \end{lstlisting}};
    \node[text width=7cm, align=center, below=-0.3cm of s2, font=\fontfamily{Roboto}\selectfont\small] (l2) {(b) Sample from Silo 2 ($S_2$) with obfuscated column names};

\end{tikzpicture}
\caption{Examples of input/output pairs of a question paired with the target SQL query which (a) are from the ChatGPT rephrased silos and (b) use column names obfuscated by an arbitrary but consistent mapping.}
\label{fig:gpt_obf_example}
\end{figure}

We limit the scope of generated SQL statements. All statements generated from our CFG, an excerpt of which is shown in the appendix, are \textsc{select} statements that only contain the SQL keywords \textsc{from}, \textsc{natural join}, and \textsc{where}. The majority of the complexity is in the \textsc{where} clause which requires specialized knowledge about the schema along with language comprehension to properly generate based on the input question. The task for the LLM is to generate the \textsc{where} clause of an SQL statement which answers the input question.

We introduce a useful normalization, for which we provide ablations in the results section. This normalization is closely related to 6NF \cite{6nf_Date2003}. It ensures that joins are natural and that column names are easily identified. In general, this transformation could help all SQL LLMs. One could provide columns with unique names, factor away complex relationships, and design redesign schemas such that joins are natural. This transformation is bidirectional, one could normalize such a schema, generate a query against it, and then transform those queries back to the original schema. Little to no work exists on database normalizations that are specific to LLMs. Existing database normalizations focus on ease of query generation for humans and concerns like execution efficiency for machines. As described in the results section, our composition methods are far superior irrespective of this normalization. But we believe it is a valuable observation that is likely to lead to many more LLM-specific normalizations as they become serious consumers of SQL.

From each SQL schema we randomly generate an instance of that SQL database. For each sample, the LLM's output is assembled into an SQL statement then executed on the database to obtain the query results, if the query results are exactly equivalent to the query result of the ground truth SQL statement then the LLM's output is considered correct. Furthermore, we ensure that each query in training and validation responds with at least one record to avoid trivial false positives. Given that natural language questions can lead to long queries, this is a high bar, as even the smallest mistake leads to zero performance.


We employ a second score that parses the generated query conditions into a tree then calculates the tree-edit distance \cite{editdist_zhang1996} between the ground query and the generated query. This is the number of edit operations required to transition between the two, which are normalized by the number of nodes in the ground tree. These are averaged across all queries for a silo or collection of silos. This edit distance score provides a far more fine-grained view into the performance of models, fine-tunings, and compositions of fine tunings.


\begin{table*}[h]
\scalebox{0.84}{
\begin{tabular}{l|rr|rrrrrr}\toprule
CFG &Baseline &Baseline & & & Ours & Ours (Logits) & \\
Generated &Exponential &Generalized & &PEM &(Maximum & Without DB & \textbf{Ours} \\
&Model &Model &LoraHub &Addition &Difference) &Normalization &\textbf{(Logits)} \\\midrule
$\text{Silos}_1$ &0.0 (100.0\%) &0.0 \phantom{0}(98.3\%) &1.9 &1.0 &0.4 &0.4 &\textbf{0.1} \\
$\text{Silos}_2$ &0.0 \phantom{0}(96.7\%) &0.0 (100.0\%) &2.7 &0.7 &0.3 &0.2 &\textbf{0.0} \\
$\text{Silos}_3$ &0.0 (100.0\%) &0.0 (100.0\%) &1.2 &0.7 &0.2 &\textbf{0.1} &\textbf{0.1} \\
$\text{Silos}_{1 \cup 2}$ &0.0 \phantom{0}(98.3\%) &1.0 \phantom{00}(0.0\%) &1.8 &1.0 &0.9 &0.7 &\textbf{0.3} \\
$\text{Silos}_{1 \cup 3}$ &0.0 \phantom{0}(99.2\%) &0.5 \phantom{00}(0.0\%) &1.6 &0.7 &0.7 &0.6 &\textbf{0.2} \\
$\text{Silos}_{2 \cup 3}$ &0.0 (100.0\%) &0.4 \phantom{00}(1.7\%) &1.7 &0.7 &0.7 &0.5 &\textbf{0.4} \\
$\text{Silos}_{1 \cup 2 \cup 3}$ &0.0 (100.0\%) &0.5 \phantom{00}(1.7\%) &2.2 &0.7 &0.7 &0.4 &\textbf{0.4} \\
\midrule $\mu\pm\sigma$ & $0.00\pm0.00$ & $0.35\pm0.35$ & $1.88\pm0.45$ & $0.78\pm0.13$ & $0.55\pm0.24$ & $0.42\pm0.21$ & $\boldsymbol{0.21\pm0.12}$ \\
\bottomrule
\end{tabular}}
\caption{Normalized tree edit distance for CFG-generated question and SQL pairs with accuracy reported in parentheses (average and std. dev. only applies to normalized tree edit distance). The exponential baseline sees all combinations of silos at training time, this is intractable and insecure, but has maximal performance. The generalization baseline sees all silos but not combinations of silos at training time, this is tractable but insecure. The other methods are used to build a SecureLLM. As described above, we do not include detailed reports on methods which underperform both LoraHub and PEM Addition. Note that our methods significantly outperform prior work. They retain all the generalization performance there is (since the generalization model sees all silos at once, while the fine-tunings each see silos separately, the generalization model should nominally perform better), even outperforming the generalization baseline. }
\label{table:exp1}
\end{table*}

\begin{table*}[h]
\scalebox{0.84}{
\begin{tabular}{l|rr|rrrrrr}\toprule
GPT &Baseline &Baseline & & & Ours & Ours (Logits) & \\
Generated &Exponential &Generalized & &PEM &(Maximum & Without DB & \textbf{Ours} \\
&Model &Model &LoraHub &Addition &Difference) &Normalization &\textbf{(Logits)} \\\midrule
$\text{Silos}_1$ &\phantom{00}0.0 (87.5\%) &0.1 (79.2\%) &2.0 &1.1 &0.5 &0.6 &\textbf{0.3} \\
$\text{Silos}_2$ &0.2 (60.8\%) &0.2 (56.7\%) &2.9 &1.0 &0.5 &\textbf{0.3} &0.4 \\
$\text{Silos}_3$ &0.1 (56.7\%) &0.2 (51.7\%) &1.4 &1.1 &0.5 &0.4 &\textbf{0.2} \\
$\text{Silos}_{1 \cup 2}$ &0.2 (20.8\%) &0.4 \phantom{0}(0.0\%) &2.1 &0.7 &0.6 &0.6 &\textbf{0.2} \\
$\text{Silos}_{1 \cup 3}$ &0.2 (29.2\%) &0.4 \phantom{0}(0.0\%) &1.6 &1.0 &0.6 &0.6 &\textbf{0.4} \\
$\text{Silos}_{2 \cup 3}$ &0.1 (33.3\%) &0.3 \phantom{0}(3.3\%) &2.3 &0.6 &0.5 &0.4 &\textbf{0.3} \\
$\text{Silos}_{1 \cup 2 \cup 3}$ &0.1 (50.0\%) &0.3 \phantom{0}(2.5\%) &2.1 &0.6 &0.5 &0.4 &\textbf{0.2} \\
\midrule $\mu\pm\sigma$ & $0.15\pm0.06$ & $0.28\pm0.13$ & $2.06\pm0.45$ & $0.85\pm0.21$ & $0.52\pm0.05$ & $0.48\pm0.11$ & $\boldsymbol{0.30\pm0.07}$ \\
\bottomrule
\end{tabular}}
\caption{Results on the ChatGPT-paraphrased questions. See \cref{table:exp1} for a detailed explanation. Our method continues to outperform all others, and again outperforms the generalization baseline. Scaling to realistic queries still favours our approach.}
\label{table:exp2}
\end{table*}

\begin{table*}[h]
\scalebox{0.82}{
\begin{tabular}{l|rr|rrrrrr}\toprule
Obfuscated &Baseline &Baseline & & & Ours & Ours (Logits) & \\
Generated &Exponential &Generalized & &PEM &(Maximum & Without DB & \textbf{Ours} \\
&Model &Model &LoraHub &Addition &Difference) &Normalization &\textbf{(Logits)} \\\midrule
$\text{Silos}_1$ &0.0 \phantom{0}(99.2\%) &0.0 \phantom{0}(94.2\%) &2 &1.1 &0.5 &0.5 &\textbf{0.2} \\
$\text{Silos}_2$ &0.0 \phantom{0}(92.5\%) &0.0 (100.0\%) &3.1 &1.4 &0.5 &\textbf{0.4} &\textbf{0.4} \\
$\text{Silos}_3$ &0.0 (100.0\%) &0.0 (100.0\%) &0.9 &0.8 &0.5 &0 &\textbf{0.1} \\
$\text{Silos}_{1 \cup 2}$ &0.0 \phantom{0}(80.8\%) &0.7 \phantom{00}(0.8\%) &1.6 &2.2 &1.1 &0.9 &\textbf{0.5} \\
$\text{Silos}_{1 \cup 3}$ &0.0 \phantom{0}(98.3\%) &0.4 \phantom{00}(0.0\%) &1.3 &1.4 &0.7 &0.6 &\textbf{0.4} \\
$\text{Silos}_{2 \cup 3}$ &0.0 \phantom{0}(77.5\%) &0.6 \phantom{00}(1.7\%) &1.6 &2.5 &0.9 &\textbf{0.7} &0.8 \\
$\text{Silos}_{1 \cup 2 \cup 3}$ &0.0 (100.0\%) &0.4 \phantom{00}(1.7\%) &1.9 &2.5 &1.0 &0.6 &\textbf{0.3} \\
\midrule $\mu\pm\sigma$ & $0.01\pm0.01$ & $0.31\pm0.28$ & $1.76\pm0.65$ & $1.70x\pm0.65$ & $0.73\pm0.24$ & $0.54\pm0.24$ & $\boldsymbol{0.36\pm0.22}$ \\
\bottomrule
\end{tabular}}
\caption{Following \cref{table:exp1} while obfuscating the table names. Our methods continue to perform well showing that they are not taking advantage of a trivial solution. For real-world applications, one would likely use a much larger baseline model. This would improve the absolute execution scores, which would method would benefit from since it retains the performance of the underlying model in this challenging compositional task.}
\label{table:exp3}
\vspace{-2ex}
\end{table*}

\section{Experiments}

To demonstrate the capabilities of model composition at inference-time, we first begin by obtaining individual fine-tunings that are knowledgeable in a single silo by fine-tuning a Llama-2-7B model for each silo separately. The fine-tuning results in a Low-Rank Adaptation (LoRA) for each silo which can independently be applied to the base Llama-2-7b model. Once the individual LoRA fine-tunings are obtained we compose them using one of several compositional schemes with the requirement that the composition happens at inference time with no additional training. We additionally train two insecure baseline models that act as an upper-bound using LoRA, the baseline generalized model is trained on all the individual silos together and must then generalize it's knowledge to the union silos. While the baseline exponential model also breaks privacy guarantees by training on all the individual silos along with the union silos, the term exponential refers to the fact that training such a model while preserving privacy would mean that $\mathcal{O}(2^N)$ models would need to be trained where $N$ is the number of silos in the database. Both baseline models are considered insecure as there is no method of removing knowledge about certain silos at inference time when the user does not have the sufficient credentials unlike our proposed SecureLLM method which can remove and add fine-tunings with each silo's knowledge at inference time. We fine-tune all models with one epoch until saturation (achieving near 100\% accuracy on the CFG validation set) using a frozen Llama-2 7B \citep{llama_touvron2023} with a trainable LoRa fine-tuning \citep{lora_hu2021} using LoRa parameters $r=8, \alpha=32$ and a dropout \citep{dropout_srivastava2014} of $0.1$, an Adam optimizer \citep{adam_2014} with a learning rate of $0.0002$, a batch size of $32$, and a weight decay of $0.002$.

We report the results of the two insecure baseline models along with the secure composition ($M_1 \oplus M_2 \oplus M_3$ where $M_i$ was trained on $\text{Silo}_i$) using multiple compositional methods (\cref{subsub:LoraHub}, \cref{subsub:MaximumDifference}) including our best method (in \cref{subsub:LogitComposition}) with and without using 6NF-like database normalization, which is equivalent to the scenario where a user has credentials to access $T=\{1, 2, 3\}$. We note that neither the baseline generalized model nor the secure compositions have seen the union Silos ($S_{1 \cup 2}$, $S_{1 \cup 3}$, $S_{2 \cup 3}$, and $S_{1 \cup 2 \cup 3}$) and that only the exponential baseline model has been trained on those silos. The performance of the composed fine-tunings on the individual Silos would give an indication as to whether the resulting composition is able to retain the knowledge of each individual fine-tuning from each separate Silo; This performance is expected to be traded off for privacy while the better compositional methods mitigate the extent of this trade off and maintain maximal privacy. The performance on the union Silos indicates whether the composed fine-tunings are able to successfully generalize knowledge from the individual fine-tunings which is an essential component in answering questions that no individual fine-tuning or silo can answer.


\section{Results}

The highest performance one can possibly achieve is if the LLM is trained not just on every silo but the powerset of silos, i.e., the insecure baseline exponential model described above. Realistically, our model is upper-bounded by a variant of the model that sees all of the silos at training time, but sees no combination of silos. Both of these are insecure, in that they have access to all of the data. Our goal is to find a method to combine individual silo fine-tunings to reproduce the performance of the baseline generalized model.

Raw overall performance is not a relevant metric, although we report it in each case for the baseline models. Raw performance is a function of the size of the model. And we use the modestly sized Llama-2-7B. What is critical is the fraction of retained performance. This is what we focus our results on, the difference in tree edit distance.

LoraHub and PEM addition were the only two competitive methods that were previously published. All other methods described earlier performed so poorly we did not include them in the final results table to make room for additional experiments.

In \cref{table:exp1} we report performance for the CFG-generated data. Note that for every probe silo combination our methods have by far the lowest tree edit distances. Even without the database normalization described above, our methods outperform all others in every case. With the database normalization our method retains all the performance that exists, i.e., it nearly always matches the tree edit distance of the baseline generalized model.

One might wonder if these results are merely an artifact of the CFG-based approach. When replicating the same experiment with sentences rephrased by ChatGPT, see \cref{table:exp2}, we come to the same conclusions. LoraHub and PEM Addition, along with all prior methods we attempted significantly underperform our approach. Note that this is an extremely challenging test set as the ChatGPT paraphrases are only used for testing, not for training.

To guard against a potential trivial solution to this problem, we also introduce a column-name obfuscated version. A model that is good at guessing a likely name for a table based on the entities it refers to might otherwise get a leg up. We are interested in the ability of models to retain compositional reasoning, rather than circumventing the task. In any case, in real world conditions column names are often rather complex. In \cref{table:exp3} each column is given an arbitrary but stable and coherent name, in this case an animal. Relative to \cref{table:exp1} our method loses little performance, meaning that it encourages compositional reasoning.

\section{Conclusion}

LLM security is critical to numerous commercial and government applications. We take a different view of LLM security compared to that of prior work, one where we import the traditional notion of access security to LLMs. This is enabled by the new compositional methods we introduce that prove themselves to be effective.

Note that our experiments are an extreme case, one where no prior knowledge is useful for understanding the structure of a silo, particularly in the case where column names are obfuscated. We also consider a weak LLM, Llama-2-7B. We showed that these novel composition methods are able to take advantage of the generalization capabilities of the LLM with SQL edit distances that are the same or even at times better than the baseline LLM when it attempts to generalize. In other words, fine-tuning the LLM on each silo jointly, performs as well as fine-tuning on each silo individually and combining the fine-tunings. This is as much as one could hope for. Practical applications would need to use a far stronger underlying LLM to achieve high execution accuracy.

There are numerous possible extensions of this work, including applications to document QA where each silo is a collection of documents rather than a database. One possible followup could look at the converse task, given a question determine the silos necessary to answer it. This could be used to monitor conversations or to automatically mark the security level of an exchange between users. Another possible direction would be to look at negative silos that exclude information. A negative silo would explicitly avoid a topic, which would prevent accidental leaks. Models could rewrite text or data to refer or exclude particular silos. The traditional world of access security is rich with problems for LLMs to address, and our work opens up the way for doing so. In addition, by providing provable security, i.e., there can be no leaks from silos the user doesn't have access to, we take a key step toward enabling the use of LLMs in secure environments.

\section{Additional Information}\label{sec:limitations}

\subsection{Limitations}

We disentangle and address a very specific slice of LLM safety, one that is often commingled with a larger story about safety.
SecureLLM only concerns itself with security in the traditional sense: quantized access permissions to data.
It relies on traditional security techniques to manage access permissions. There is a widely
held belief that LLM security is a totally disjoint new field, but as we show, with the SecureLLM
approach we can reduce many of those security problems back to traditional access permission issues.
Many security problems of LLM are manageable through traditional means when one can assume that
only vetted actors have access, just as they are with current document storage systems. The same
systems which ensure patient privacy, financial privacy, and that manage secret information today can be
used to manage collections of fine-tunings, and the same supervision methods can trace access to the LLM.

From this perspective SecureLLM solves data leakage and prompt injection attacks, in the same sense that
traditional security solves data leakage: those without permissions cannot access this information,
and those with permissions have full access with training and supervision. Although, in the future one might imagine extensions that
provide more fine-grained permissions. Organizations are already set up for this form of security, 
both for managing the user permissions and for the associated documents, making the deployment of SecureLLM
straightforward. In settings where this structure is not available or not appropriate, SecureLLM is
not directly applicable, but those scenarios never had meaningful security to begin with.

Explicitly out of scope are other notions of safety and security. For example, the LLM may still
fabricate information, produce toxic or biased results, follow guidance that it should not, etc.
The only mitigation that we offer is that only a user that has permissions to that data will
be impacted directly; hopefully this user will receive appropriate training about the limitations
and dangers of LLMs.

\subsection{Ethics}

SecureLLM could pose some ethical issues.
At the moment, surveillance is limited by the need to process a deluge of data. This results in mostly processing metadata. Enabling LLMs to work in secure environments could contribute to large-scale monitoring, detection, and tracking. Novel uses of LLMs in secure environments can advance numerous defense applications.
As with many other dual-use technologies, we hope that this work will be used for positive ends.

\subsection{Computational Considerations}

We use NVIDIA Titan RTX 24GB VRAM GPUs for all our experiments.

For all of our PEFT parameters, less than one GPU-hours per PEFT was required for training. Between around one GPU-hours was required for composition growing with respect to the number of compositions. For each run, approximately 20 GB of VRAM is needed as we use half precision for all training and inference. Each PEFT can be trained and inferenced on one GPU.

We estimate a total of 10 GPU-hours is required to replicate results for training, and 20 GPU-hours is required to perform the same experiments described in this manuscript.

\subsection{Reproducibility}

All code and data required to reproduce our work will be provided at this GitHub repository under the MIT license \url{https://github.com/Scuwr/SecureLLM}.

\subsection{Acknowledgements}

This work was supported by the DARPA Machine Common Sense (MCS) program, the Center for Brains, Minds, and Machines, NSF STC award CCF1231216, the NSF award 2124052, the MIT CSAIL Machine Learning Applications Initiative, the
MIT-IBM Watson AI Lab, the DARPA Artificial Social Intelligence for Successful Teams (ASIST)
program, the DARPA Knowledge Management at Scale and Speed (KMASS) program, the United
States Air Force Research Laboratory and the Department of the Air Force Artificial Intelligence
Accelerator under Cooperative Agreement Number FA8750-19-2-1000, the Air Force Office of
Scientific Research (AFOSR) under award number FA9550-21-1-0014, and the Office of Naval
Research under award number N00014-20-1-2589 and award number N00014- 20-1-2643. The views
and conclusions contained in this document are those of the authors and should not be interpreted as
representing the official policies, either expressed or implied, of the Department of the Air Force or
the U.S. Government. The U.S. Government is authorized to reproduce and distribute reprints for
Government purposes notwithstanding any copyright notation herein.

\bibliography{anthology,custom}
\bibliographystyle{acl_natbib}

\newpage
\onecolumn
\appendix



\section{Negative Results}
\label{sec:appendix_negative}

We briefly outline some methods that seemed promising but did not perform well, far worse than PEM Addition in testing. 

\textbf{Strongest Output Embedding Response} Only using the LoRA Layer, $L_i$, that generated the strongest output embedding response at a given attention layer produced compositions that were 50\% less effective than PEM Addition in initial informal tests.

\textbf{Element-wise Maximum} Using the standard max function in pytorch, $max\left([L_1(x), \cdots, L_n(x)], dim=0\right)$, produced compositions that were 50\% less effective than PEM Addition in initial informal tests.

\section{An example of the Context-Free Grammar (CFG)}

We provide an excerpt from the CFG of one of the silos in \cref{fig:cfg} which is parsed by our custom parser as to jointly construct a natural language questions alongside an SQL query.
\label{sec:appendix_cfg}
\begin{figure*}[h]
\centering
\begin{lstlisting}

:1:root:{whats_the} average {prop_number} {suffix}:T=AVG
:1:root:{whats_the} maximum {prop_number} {suffix}:T=MAX
:1:root:{whats_the} minimum {prop_number} {suffix}:T=MIN
:1:root:{list_the} {prop_number} {suffix}:T=none
:1:root:{whats_the} number of {prop_countable} {suffix}:T=COUNT
:1:root:{list_the} names of all teachers {teacher_multi_cond} {suffix}:T=none ; WANT=instructors.name ; TABLES = instructors
# {more not shown here}
:1:list_the:output a list of the:none
:1:list_the:provide a list of the:none
:1:whats_the:what is the:none
:1:whats_the:what's the:none
:1:suffix:in our records:none
:1:suffix:in the school:none
# {more not shown here}

:1:prop_countable:teachers {teacher_multi_cond}:TABLES = instructors
:1:prop_countable:{class}:TABLES = classes
:1:prop_number:age of all teachers {teacher_multi_cond}:WANT=instructors.teacher_age ; TABLES = instructors
:1:prop_number:grade of all {class}:WANT=classes.grade ; TABLES = classes
# {more not shown here}
:1:teacher_multi_cond::none
:3:teacher_multi_cond:{teacher_cond_taught}:none
:2:teacher_multi_cond:{teacher_cond_age}:none
:4:teacher_multi_cond:{teacher_cond_age} and {teacher_cond_taught}:none
:4:teacher_multi_cond:{teacher_cond_age} or {teacher_cond_taught}:ADDER_PROPS={"comb":"OR"}
:1:teacher_cond_taught:that taught {class}:TABLES = classes
:1:teacher_cond_age:that are older than {number,20,90}:CONDITIONS=instructors.teacher_age >= {0}
# {more not shown here}

:1:class:{subject} classes:none
:2:class:{subject} classes that {class_avg}:none
:2:class:{subject} classes that {class_avg} and {class_year}:none
:2:class:{subject} classes {class_level_single}:none
:2:class:{subject} classes {class_level_single} that {class_year}:none
:1:class_avg:achieved a grade higher than {number,60,95}:CONDITIONS=classes.grade >= {0}
:1:class_year:were conducted before {number,2000,2020}:CONDITIONS=classes.year <= {0}
# {more not shown here}

:1:subject::none
:1:subject:{single_subject}:CONDITIONS=classes.class_subject = '{0}'
:1:subject:{single_subject} or {single_subject}:CONDITIONS=classes.class_subject = '{0}' OR classes.class_subject = '{1}'
# {more not shown here}
:1:single_subject:math
:1:single_subject:economics
:1:single_subject:history
# {more not shown here}
\end{lstlisting}
\caption{An excerpt of the Context-Free Grammar for Silo 1 ($S_1$). The content is shortened with many lines removed as to fit inside the figure while conveying the general structure of the grammar. Each rule contains a pair of a natural language phrase followed by a colon and the sufficient information required to properly build the resulting SQL which is our own custom }
\label{fig:cfg}
\end{figure*}

\end{document}